\documentclass[conference]{IEEEtran}

\usepackage[T1]{fontenc}
\usepackage[utf8]{inputenc}
\usepackage{cite}
\usepackage{amsmath,amssymb}
\usepackage{array}
\usepackage{booktabs}
\usepackage{graphicx}
\usepackage[table]{xcolor}
\usepackage[colorlinks,urlcolor=blue,linkcolor=blue,citecolor=blue]{hyperref}

\begin{document}

\title{Scene-Consistent Illumination Transfer for Inserted Advertising Graphics}

\author{
\IEEEauthorblockN{Rameshwar Mishra\IEEEauthorrefmark{1}, Bishshoy Das\IEEEauthorrefmark{2}, A. V. Subramanyam\IEEEauthorrefmark{1}, Guan-Ming Su\IEEEauthorrefmark{2}}
\IEEEauthorblockA{\IEEEauthorrefmark{1}Indraprastha Institute of Information Technology, Delhi, India\\
\{rameshwarm, subramanyam\}@iiitd.ac.in}
\IEEEauthorblockA{\IEEEauthorrefmark{2}Dolby Laboratories, Bangalore, India and Sunnyvale, USA\\
\{bishshoy.das, guanming.su\}@dolby.com}
}

\maketitle

\begin{abstract}
Replacing a visible advertisement in a broadcast frame is geometrically straightforward but photometrically delicate. A pasted graphic can have the correct perspective and still appear detached when its brightness, shading, or shadow disagrees with the surface beneath it. This paper presents \emph{Ad-Relight}, an inference-only procedure for transferring scene illumination to a supplied advertising graphic without collecting a banner-specific training set. The procedure first separates slowly varying shade from graphic structure, then probes a pretrained diffusion relighter with two nearly identical backgrounds to isolate the contribution of the target region. A final pass combines this residual with a smoothed luminance field and a soft attenuation mask. Across 560 generated placements, the approach improves structural similarity, perceptual distance, and illumination agreement over geometric compositing and direct relighting baselines. Human judgments and an automated preference study show the clearest gains on floor-mounted graphics with nonuniform lighting. The current study is image based; temporal stabilization remains an open extension.
\end{abstract}


\section{Task and Contributions}

\begin{figure}[t]
    \centering
    \includegraphics[width=\columnwidth]{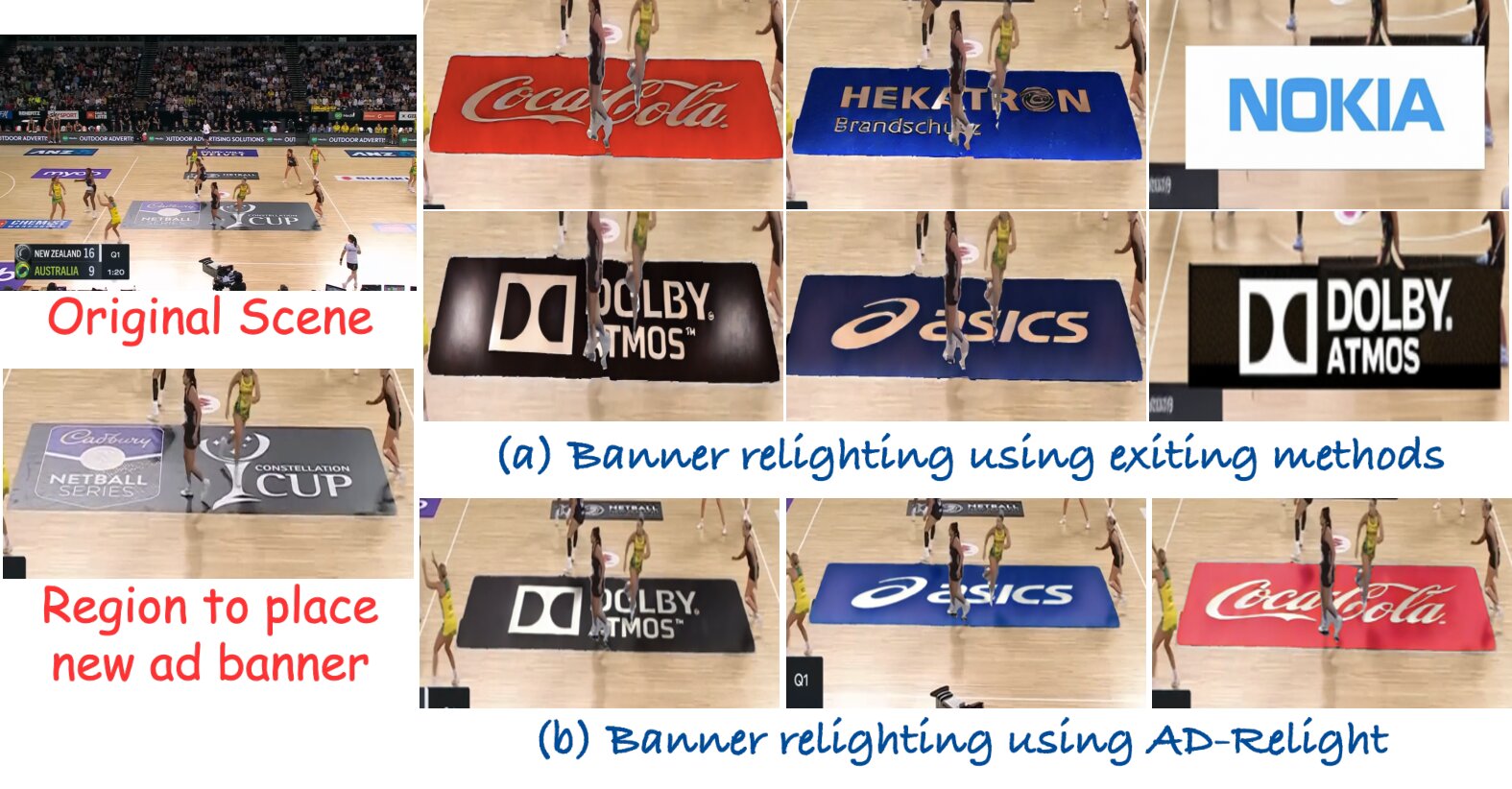}
    \caption{A supplied graphic can be geometrically correct yet visually detached from its host surface. The examples motivate matching spatial brightness changes, rather than treating the replacement as a uniformly lit texture.}
    \label{fig:teaser}
\end{figure}

Personalized advertising often begins with a finished graphic and a location in an existing frame. The production system must preserve the graphic's identity while making it look as though it was present when the scene was captured. This requirement is especially visible on courts, floors, walls, and other broad surfaces: the replacement area may contain a brightness ramp, a soft occlusion, or a color cast that cannot be reproduced by a homography alone. Figure~\ref{fig:teaser} illustrates the resulting gap between geometric placement and scene-aware appearance.

Let $I$ denote a frame, $M$ a mask for the region to be replaced, and $L$ a user-supplied banner. The desired output should satisfy three constraints: its boundary must follow $M$, the semantic content of $L$ must remain legible, and its low-frequency appearance should agree with the illumination already present in $I$. These constraints pull in different directions. Strong generative editing can improve realism while changing a logo, whereas a literal paste preserves the logo while exposing the mismatch in lighting.

Our design treats a pretrained relighting network as a source of illumination evidence rather than as a direct banner generator. A pair of controlled queries reveals how the network responds to the original target region; that response is then reused while the supplied graphic remains explicit throughout the pipeline. This separation makes the method usable at inference time and avoids fitting a new relighting model to a small, specialized advertising collection.

The paper makes three contributions:
\begin{itemize}
    \item a formulation of banner replacement as region-conditioned illumination transfer, with identity preservation treated as a first-class constraint;
    \item a training-free procedure that combines shade normalization, differential probing, and soft shadow attenuation under the name \emph{Ad-Relight};
    \item a focused evaluation covering 560 placements, component ablations, automated rankings, and human pairwise preferences.
\end{itemize}

\section{Positioning}

Classical compositing methods modify image gradients or local appearance to hide a seam. Poisson editing is a representative example: it provides a principled way to blend boundaries, but it does not by itself infer the illumination that should fall across a new planar object~\cite{perez2003poisson}. Learned harmonization systems use broader image context to align a foreground with its background~\cite{tsai2017harmonization}; they are effective when the training distribution covers the desired composite, yet a printed floor banner has a distinctive combination of planar perspective, texture, and spatial light variation.

Diffusion models provide a useful alternative because their denoising process contains broad visual priors~\cite{ho2020ddpm}. Recent illumination-editing work imposes light-transport consistency during training and exposes a strong general-purpose relighting backbone~\cite{zhang2025iclight}. Directly applying such a backbone to a pasted banner is not sufficient: the network may interpret a horizontal graphic as part of the floor, or may alter the lettering while attempting to improve realism. Ad-Relight therefore uses the network twice for diagnosis and once for synthesis. The central distinction is not a new generative model, but a test-time construction that extracts a local lighting signal before asking the model to edit the banner.

\section{Ad-Relight}

\begin{figure*}[t]
    \centering
    \includegraphics[width=\textwidth,height=0.235\textheight,keepaspectratio]{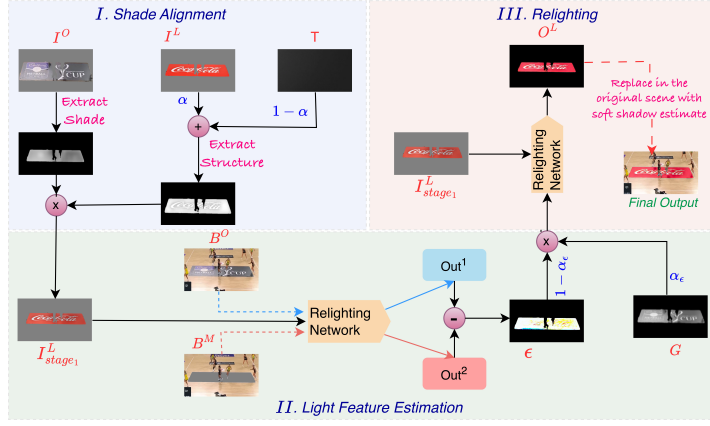}
    \caption{Processing graph for Ad-Relight. A low-frequency shade field is transferred to the supplied graphic, two backbone queries provide a target-region residual, and a mixture of that residual with a smoothed luminance guide controls the final relighting pass.}
    \label{fig:pipeline}
\end{figure*}

Figure~\ref{fig:pipeline} summarizes the pipeline. All operations are performed independently for one frame. The relighting backbone is frozen, and no banner examples are used to update its parameters.

\subsection{Region-conditioned preparation}

We crop the target support using $M$ and denote its original appearance by $O$. The replacement graphic is first placed in the same planar coordinate system; this geometric step can be obtained from the host application's mask or a homography. To avoid a perfectly synthetic surface, a mild texture layer $T$ is mixed into the graphic $L$:
\begin{equation}
    L_{\mathrm{tex}}=(1-\alpha)L+\alpha T,
    \qquad 0\leq\alpha\leq1 .
\end{equation}
The texture weight is fixed at $\alpha=0.3$ in our experiments. It is deliberately small: texture should support contact with the scene without competing with the supplied branding.

The main preparation step transfers only slowly varying shade. Let $Y_O$ and $Y_L$ be the luminance channels of $O$ and $L_{\mathrm{tex}}$, respectively, and let $\mathcal{G}_K$ denote a Gaussian low-pass operator. We form a smooth component and a normalized structural component for each image:
\begin{equation}
\begin{aligned}
S_O&=\mathcal{G}_K(Y_O), & R_O&=\frac{Y_O}{S_O+\delta},\\
S_L&=\mathcal{G}_K(Y_L), & R_L&=\frac{Y_L}{S_L+\delta}.
\end{aligned}
\end{equation}
Here $\delta$ prevents unstable division in dark pixels. The banner's luminance is replaced by $Y_L^{\star}=S_OR_L$. Thus, the graphic retains its internal contrast while inheriting the broad shade pattern measured on the host surface. This operation is intentionally conservative: it does not ask a generator to redraw letters, edges, or colors.

\subsection{Differential illumination probe}

The prepared banner is not yet a reliable conditioning image for a relighter. We therefore query the frozen backbone with two versions of the scene. In the first query, $B_O$ contains the complete frame and the original target region. In the second, $B_M$ has the target region suppressed while the rest of the frame is unchanged. Let $F_\phi$ be the relighting network and let $Q_O$ and $Q_M$ be its corresponding responses. Under the local linear behavior encouraged by consistent-light training, the two responses can be viewed as
\begin{equation}
Q_O\approx T_\phi L_\phi,\qquad
Q_M\approx T_\phi(L_\phi-L_O),
\end{equation}
where $L_O$ is the illumination contribution associated with the target region. Their difference gives a spatially registered residual:
\begin{equation}
\epsilon=Q_O-Q_M\approx T_\phi L_O .
\end{equation}
The residual is not treated as a physically calibrated light map. It is a model-space measurement that retains the direction and relative strength of the target area's contribution. Subtraction also suppresses scene content shared by both queries, which is useful when the frame contains spectators, court markings, or other high-contrast details.

\subsection{Guided relighting and contact attenuation}

The residual can contain small traces of appearance leakage. We stabilize it with a second guide obtained from the target luminance:
\begin{equation}
G=\mathcal{G}_{K'}(Y_O),\qquad
B_{\epsilon}=\alpha_{\epsilon}G+(1-\alpha_{\epsilon})\epsilon,
\end{equation}
where $K'<K$. The final backbone call uses $B_{\epsilon}$ as its lighting condition and $L^{\star}$ as the image to be edited:
\begin{equation}
O_L=F_\phi(B_{\epsilon},L^{\star}).
\end{equation}
The smaller blur radius preserves broad gradients while attenuating unstable, high-frequency differences.

Finally, we estimate contact darkening from the smoothed target luminance. Otsu thresholding supplies $\tau$, and the continuous attenuation field is
\begin{equation}
d(x,y)=\min\left(1,\frac{Y_O(x,y)}{\tau}\right).
\end{equation}
The luminance of the relit banner is mixed with its attenuated version using $\alpha_s$:
\begin{equation}
Y_{\mathrm{final}}=\alpha_sY_{O_L}+(1-\alpha_s)Y_{O_L}d.
\end{equation}
This soft weighting avoids the hard contour that a binary shadow mask would produce. We use $K=99$, $K'=21$, $\alpha_{\epsilon}=0.4$, and $\alpha_s=0.2$.

\section{Empirical Study}

\subsection{Protocol and measurements}

We evaluate the frame-level problem on 80 source frames and seven replacement graphics, producing 560 source--banner cases. The collection varies camera elevation, surface material, banner color, and the strength of the illumination gradient. The difficult subset contains large horizontal placements where the host floor is glossy or unevenly lit. Each method receives the same planar placement and the same source graphic.

We compare against four alternatives: a perspective warp with no appearance correction, a training-free image compositor, direct use of the relighting backbone, and a shadow-conditioned relighting pipeline. The last three are intended to separate the value of generic diffusion composition, an unmodified relighter, and explicit shadow guidance. Scores are computed inside the replacement mask. SSIM measures structural agreement, LPIPS measures learned perceptual distance, and ILL-SIM is the cosine similarity between the host and output luminance fields.

\begin{table}[t]
\caption{Frame-level comparison on 560 placements. Higher is preferred for SSIM and ILL-SIM; lower is preferred for LPIPS.}
\label{tab:main}
\centering
\footnotesize
\setlength{\tabcolsep}{3.2pt}
\begin{tabular}{lccc}
\toprule
Method & SSIM $\uparrow$ & LPIPS $\downarrow$ & ILL-SIM $\uparrow$\\
\midrule
Warp-only & 0.89 & 0.12 & 0.82\\
Training-free compositor & 0.45 & 0.57 & 0.62\\
Direct relighting & 0.91 & 0.07 & 0.77\\
Shadow-guided relighting & 0.92 & 0.09 & 0.79\\
\textbf{Ad-Relight} & \textbf{0.95} & \textbf{0.03} & \textbf{0.92}\\
\bottomrule
\end{tabular}
\end{table}

\subsection{Appearance comparison}

\begin{figure*}[t]
    \centering
    \includegraphics[width=\textwidth]{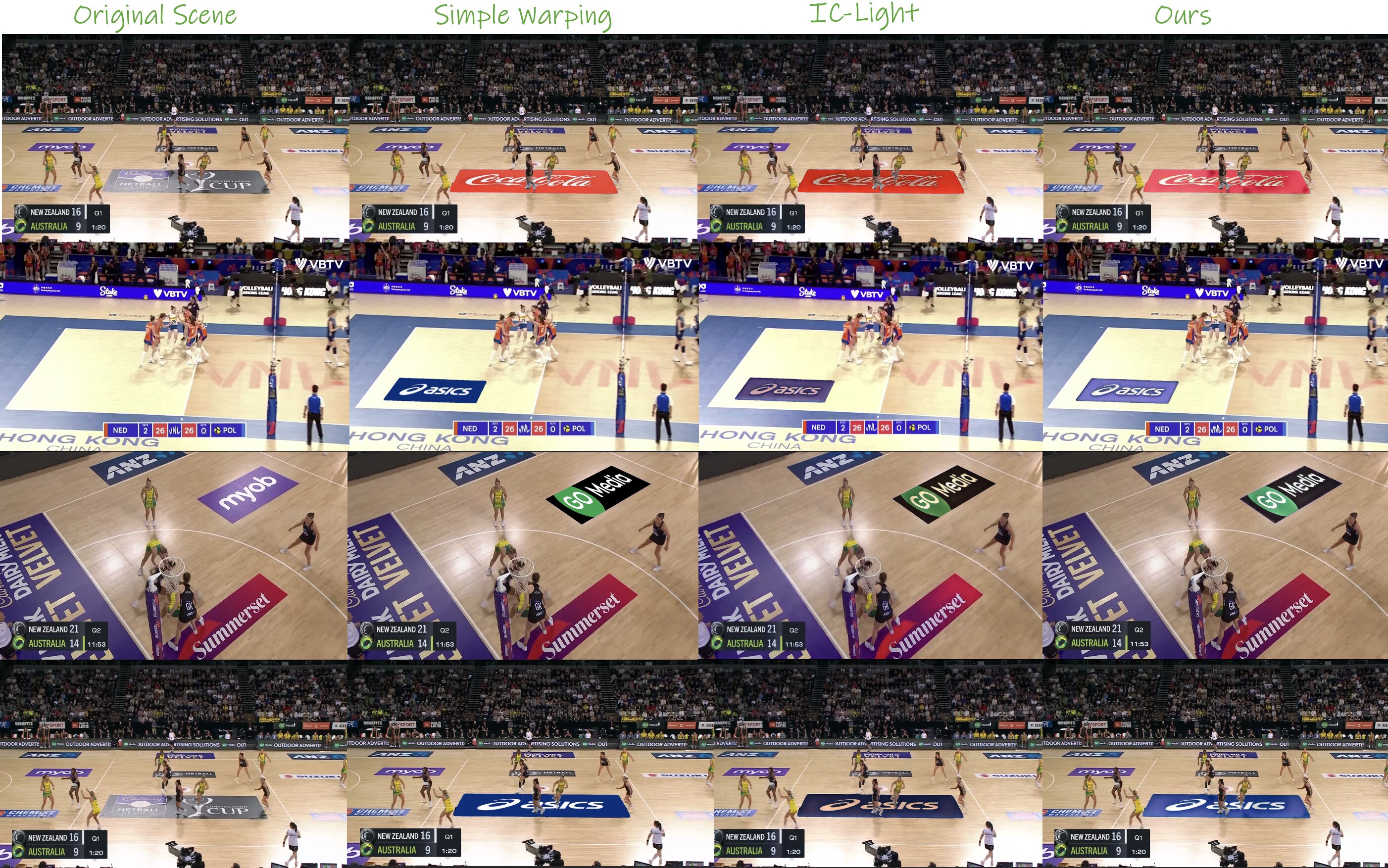}
    \caption{Representative replacements arranged by scene and method. Geometric pasting leaves the graphic uniformly lit, while direct relighting can borrow texture from the floor. The proposed pipeline preserves the logo and follows the host surface's broad brightness changes.}
    \label{fig:qualitative}
\end{figure*}

The aggregate scores in Table~\ref{tab:main} show that preserving the supplied graphic and matching the host luminance are compatible objectives. The warp-only system keeps the banner recognizable but leaves it visually flat. Direct relighting improves global appearance in ordinary cases, yet it can absorb the banner into the floor texture when the target is horizontal. The differential signal gives Ad-Relight a more localized condition, which explains its improvement in ILL-SIM as well as its lower LPIPS. Figure~\ref{fig:qualitative} shows representative sports frames with different camera heights and floor materials.

\subsection{Component sensitivity}

We test two nearby parameter settings and three targeted removals. M1 and M2 perturb the texture blend, blur radii, residual mixture, and shadow weight together. M3 removes the low-frequency guide $G$, M4 omits the shade-transfer stage, and M5 removes the differential residual $\epsilon$. The numbers in Table~\ref{tab:ablation} indicate that the residual is particularly important for perceptual similarity, while shade preparation has a strong effect on structural and illumination scores. The full configuration remains the most balanced choice.

\begin{table}[t]
\caption{Ablation on the same benchmark.}
\label{tab:ablation}
\centering
\footnotesize
\setlength{\tabcolsep}{4pt}
\begin{tabular}{lccc}
\toprule
Variant & SSIM $\uparrow$ & LPIPS $\downarrow$ & ILL-SIM $\uparrow$\\
\midrule
M1 & 0.92 & 0.03 & 0.85\\
M2 & 0.91 & 0.05 & 0.87\\
M3 ($-G$) & 0.89 & 0.07 & 0.83\\
M4 ($-$shade) & 0.86 & 0.05 & 0.80\\
M5 ($-\epsilon$) & 0.80 & 0.10 & 0.85\\
\textbf{Full} & \textbf{0.95} & \textbf{0.03} & \textbf{0.92}\\
\bottomrule
\end{tabular}
\end{table}

\begin{figure}[t]
    \centering
    \includegraphics[width=\columnwidth]{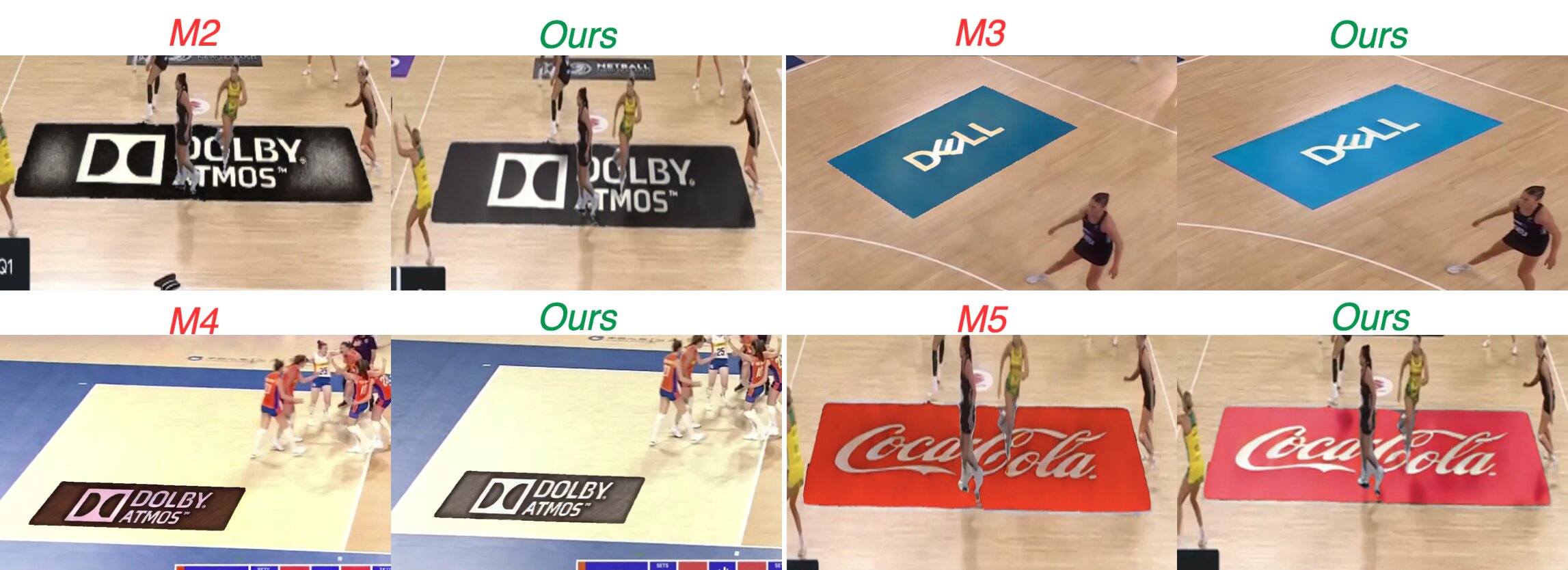}
    \caption{Ablation examples. Removing a stage changes either the banner's contact with the floor or the strength of the recovered illumination gradient.}
    \label{fig:ablation}
\end{figure}

\subsection{Preference studies}

We use two complementary preference tests. First, GPT-4o is asked to choose between unlabeled outputs using three criteria: gradient fidelity, lighting agreement, and overall realism. The percentage of cases favoring Ad-Relight is shown in Table~\ref{tab:preference}. The model favors the proposed output in every comparison, with the largest margins for the geometric baseline.

\begin{table}[t]
\caption{Percentage of GPT-4o pairwise choices favoring Ad-Relight.}
\label{tab:preference}
\centering
\footnotesize
\setlength{\tabcolsep}{3.2pt}
\begin{tabular}{lccc}
\toprule
Criterion & Warp & Direct & Shadow-guided\\
\midrule
Gradient fidelity & 100.0 & 97.6 & 95.2\\
Lighting agreement & 97.6 & 95.2 & 92.8\\
Scene realism & 90.4 & 92.8 & 85.7\\
\bottomrule
\end{tabular}
\end{table}

\begin{figure}[t]
    \centering
    \includegraphics[width=\columnwidth]{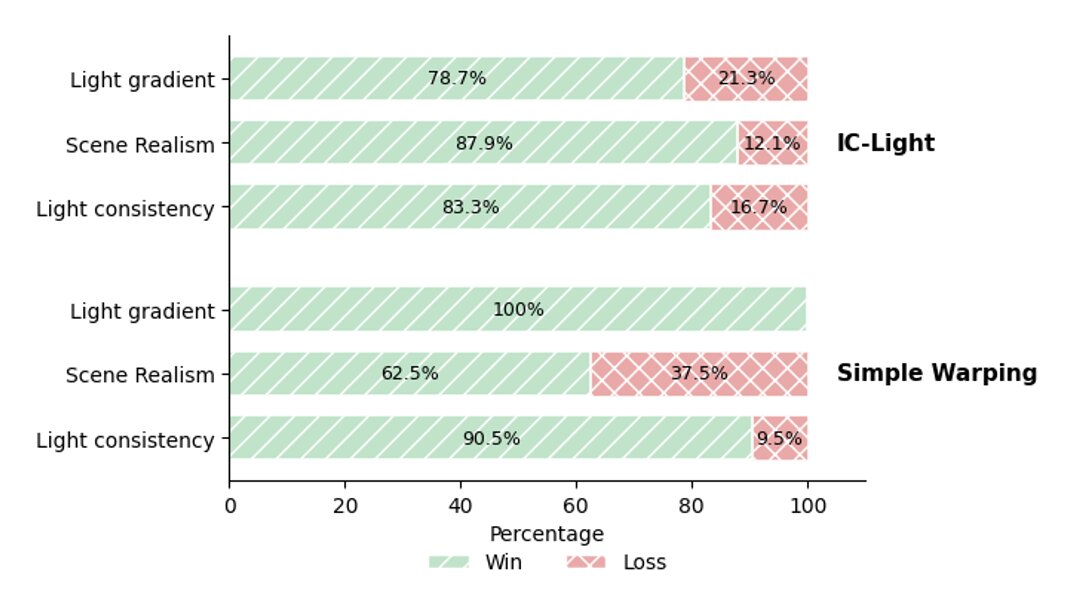}
    \caption{Human preference rates for two comparisons. Green bars indicate selections of Ad-Relight.}
    \label{fig:userstudy}
\end{figure}

Second, 25 participants completed 324 two-choice judgments across four questionnaire variants. The options were unlabeled and the task emphasized the same three properties. Figure~\ref{fig:userstudy} summarizes the preference split for the direct and warp-only comparisons. Human responses favor the proposed output most consistently when a strong gradient is present; this aligns with the quantitative ILL-SIM improvement rather than merely reflecting a preference for a particular logo color.

\subsection{Scope and failure modes}

The evidence is limited to independent frames. The approach assumes that the backbone's response changes smoothly enough for subtraction to reveal a useful regional cue; scenes with strong reflections, moving shadows, or severe occlusion can violate that assumption. The pipeline also inherits the mask and homography supplied by the placement system, so an inaccurate target boundary is not corrected by relighting. These limitations suggest that a future video version should estimate a temporally filtered residual and jointly refine geometry and illumination.

\section{Perspective}

Ad-Relight reframes banner integration as a measurement problem. Instead of asking a general-purpose generator to invent a plausible appearance for a logo, it first extracts how the host region influences a frozen relighting model and then applies that evidence to the unchanged graphic. The resulting procedure is compact, training-free, and particularly effective for planar surfaces whose lighting varies across the banner. The reported gains do not remove the need for temporal modeling or better placement masks, but they show that a carefully designed test-time probe can turn a general illumination prior into a practical tool for personalized advertising.

\end{document}